\documentclass[11pt]{article}
\usepackage{nodalida2025}
\usepackage{times}
\usepackage{hyperref}
\usepackage{url}
\usepackage{latexsym}
\usepackage{graphicx}
\usepackage{multirow}
\usepackage{subfig}

\aclfinalcopy 
\title{The Veln(ia)s is in the Details: Evaluating LLM Judgment on Latvian and Lithuanian Short Answer Matching}

\author{Yevhen Kostiuk$^{1,2}$\\
$^1$ARG-tech, \\
University of Dundee \\
$^2$OpenBabylon \\
{\tt ykostiuk001@dundee.ac.uk} \\\And
  Oxana Vitman \\
  University of Bremen  \\\And
  \L ukasz Gaga\l a \\
  Georg-August \\ Universität Göttingen \\\And
  Artur Kiulian \\
  OpenBabylon}

\date{}

\begin{document}
\maketitle
\begin{abstract}

In this work, we address the challenge of evaluating large language models (LLMs) on the short answer matching task for Latvian and Lithuanian languages. We introduce novel datasets consisting of 502 Latvian and 690 Lithuanian question-answer pairs.
For each question-answer pair, we generated matched and non-matched answers using a set of alteration rules specifically designed to introduce small but meaningful changes in the text. These generated answers serve as test cases to assess the ability of LLMs to detect subtle differences in matching of the original answers. A subset of the datasets was manually verified for quality and accuracy. Our results show that while larger LLMs, such as QWEN2.5 72b and LLaMa3.1 70b, demonstrate near-perfect performance in distinguishing matched and non-matched answers, smaller models show more variance. For instance, LLaMa3.1 8b and EuroLLM 9b benefited from few-shot examples, while Mistral Nemo 12b underperformed on detection of subtle text alteration, particularly in Lithuanian, even with additional examples. QWEN2.5 7b and Mistral 7b were able to obtain a strong and comparable performance to the larger 70b models in zero and few shot experiments. Moreover, the performance of Mistral 7b was weaker in few shot experiments.  
The code and the dataset are available on our GitHub\footnote{HIDDEN FOR REVIEW}.
\end{abstract}

\section{Introduction}

In educational domain, open-ended questions are commonly used and can be defined as questions that require a more elaborate response than simple yes-no or selection of a correct choice. These questions help to encourage a discussion, share ideas and provide more freedom for a student.

Evaluation of responses to the open-ended question is a time-consuming and difficult task that requires an evaluator to carefully read each answer and compare it with the correct answers, ensuring they match. Automating this process makes it easier for evaluators to provide a feedback and analyze errors faster~\cite{pillai2018combined, sreevidhya2021short}.

The automatic short answer matching task addresses this challenge. The goal of the task is to predict whether an answer to the question is matching a correct answer. With the introduction of LLMs, reasonable performance was achieved on English and other high-resource languages for this problem~\cite{ivanova2024evaluating}. On the other hand, when it comes to low-resource settings, LLMs demonstrated weaker results, as well as displayed biases~\cite{hackl2023gpt, lai2023chatgpt}.

In this work, we focus on Latvian and Lithuanian answer matching task, specifically on a detection of correct and incorrect responses that are similar to a set of reference ``gold'' answers, but differ in the key detail(s) to the question. 

\begin{figure}
    \centering
    \includegraphics[width=\linewidth]{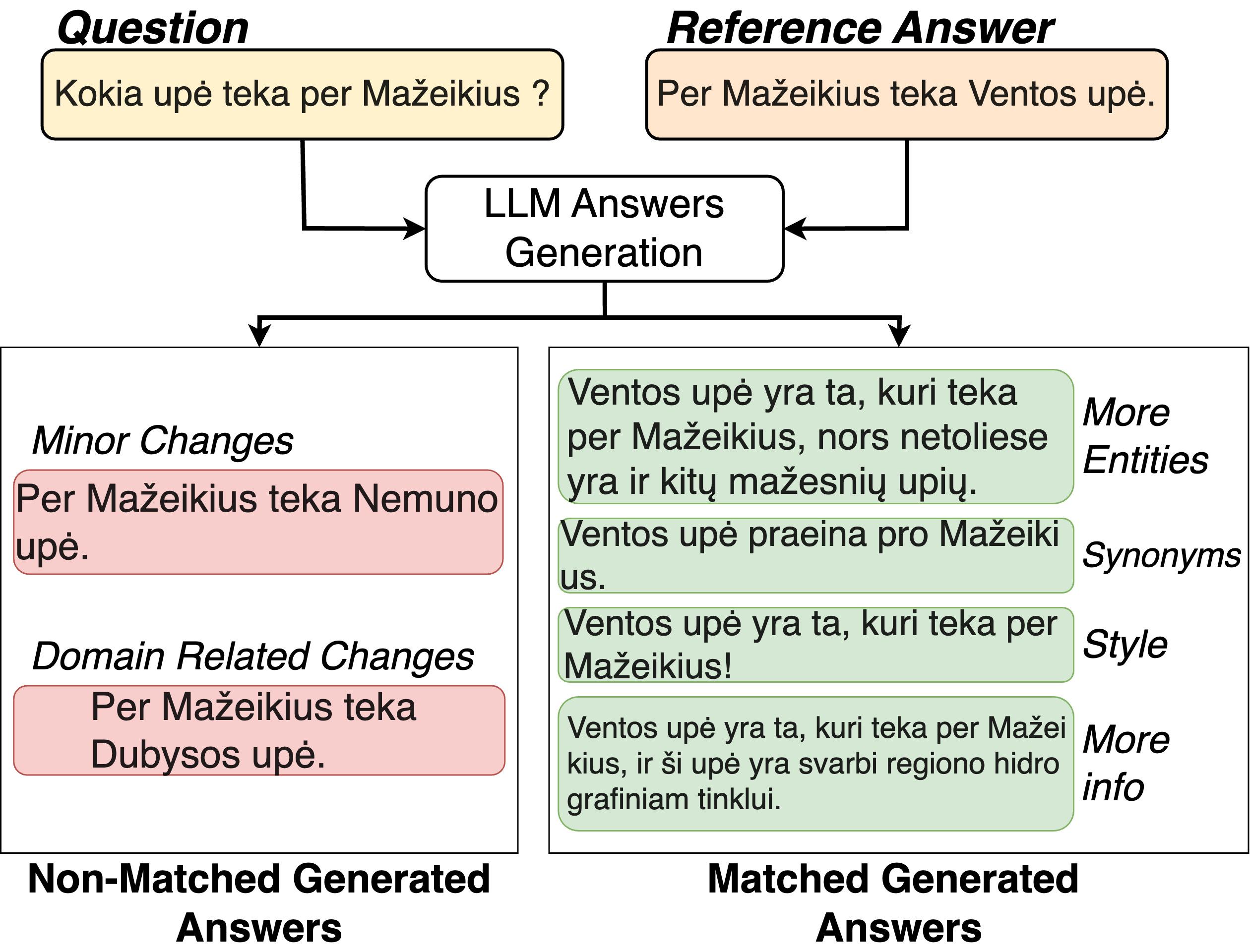}
    \caption{Example of the element from Lithuanian generated dataset.}
    \label{fig:ds_example}
\end{figure}

We automatically generated open-ended question-answer datasets for these languages based on Wikipedia. For this task, we do not focus on the factual correctness of the answers, as it does not affect the task. Each element of the dataset contains a question and its reference answer. Then we generated a set of answers that are matched with the reference answer and a set of non-matched answers. The non-matched answers are created as similarly as possible to the reference answers with respect to the words inclusion, but with the key words changed to make it incorrect. To generate the answers, we formulated different text \textit{alteration rules} (\textbf{AR}) that are minor when it comes to a text change, but semantically are major. For each rule, the different LLMs with a few shot generation process were used. Finally, to ensure the quality, we manually evaluated a sample of the data and filtered the final dataset based on it. We expect the models to obtain high, almost perfect, results on this task.

We formulated the following research questions in this paper. 
\paragraph{Q1:} Are LLMs capable of correctly identifying matched and non-matched answers with the proposed alteration rules?
\paragraph{Q2:} Is there a difference between few-shot and zero-shot inference for different LLMs for this task?

Our contributions are the following:
\begin{itemize}
    \item We automatically generated a dataset of 502 Latvian and 690 Lithuanian question-answer pairs based on Wikipedia. We defined and generated a list of matched and non-matched answers to each pair of question-answer, resulting in 3,012 and 4,830 elements for Latvian and Lithuanian respectively, and partially manually evaluated samples of the datasets.
    \item We evaluated LLaMa3.1 (8b and 70b)~\citep{dubey2024llama3herdmodels}, Mistral Nemo 12b and Mistral 7b~\citep{jiang2023mistral7b}, EuroLLM 9b~\cite{martins2024eurollmmultilinguallanguagemodels}, and QWEN2.5 (7b and 72b)~\cite{qwen2.5, qwen2} models and compared their achieved accuracy scores per AR and overall.
    \item We evaluated the models in zero-shot and few-shot settings and their performance based on different ARs of matched and non-matched answers.  
\end{itemize}

Our findings showed that larger LLMs, such as QWEN2.5 72b and LLaMa3.1 70b, consistently performed well across both Latvian and Lithuanian datasets, effectively distinguishing matched and non-matched answers in both zero shot and few shot experiments. However, smaller models demonstrated variation in their results. LLaMa3.1 8b and EuroLLM 9b showed improved performance with few-shot examples, while Mistral Nemo 12b showed limitations, particularly in Lithuanian. QWEN2.5 7b and Mistral 7b were able to obtain a similar the performance to the larger 70b models, with Mistral 7b showing weaker performance in few shot experiments.

\section{Related Work}

Answer matching task can be viewed as a subtask of the automatic short answer grading (ASAG). The definition of what is a short answer and if it is acceptable can vary depending on the domain~\cite{burrows2015eras, bonthu2021automated}. Nevertheless, all the definitions involve high semantic similarity between the correct answer(s) and predicted answers. The grading scale is also can be domain dependent~\cite{zhang2205automatic, divya2023automation, krithika2015learning}.

With the development of deep learning methods, they were widely used for the task, as they provide better robustness towards syntactic changes of the text rather than other methods~\cite{bonthu2021automated}, utilizing RNNs~\cite{cai2019automatic}, CNNs~\cite{chen2019research}, transformers~\cite{sung2019pre, willms2022transformer} and so on. Some of the suggested methods are aimed to not only grade an answer, but to explain its flows and inaccuracies~\cite{tornqvist2023exasag}. 

With the rise of generative large language models (LLMs), they were applied for ASAG as well~\cite{asag-tune, ivanova2024evaluating, chu2024llm, schneider2023towards, grevisse2024llm, yancey2023rating, yoon2023short}. Analysis of LLMs for this task showed that they are capable of predicting consistent ratings for English~\cite{hackl2023gpt, mizumoto2023exploring}. However, studies showed that the LLMs' performance on the non-English datasets is weaker~\cite{lai2023chatgpt, dargis-etal-2024-evaluating}. 

On the other hand, as any other NLP task, there is a gap in the ASAG resources for low-resource languages, including Nordic and Baltic. This area lacks high-quality datasets for these languages. The GPT-3.5 and GPT-4 models were evaluated on Finnish ASAG~\cite{chang2024automatic} on the dataset of students' answers in Finnish for multiple subjects. The study demonstrated that the models assigned higher scores to the students' answers than the human annotator and achieving Quadratic Weighted Kappa (QWK) score of 0.44. In~\cite{chang-etal-2022-towards}, the authors considered ASAG task as a paraphrase retrieval task, evaluating classical methods (TF-IDF) and different transformer methods.

In~\cite{dargis2022corpus}, the self-assessment platform for Latvian language learners was proposed and developed. The authors generated exercises automatically based on data from multiple corpora~\cite{20.500.12574/84, dargis-etal-2022-lava}. In~\cite{stefanovivc2024machine}, the research on detecting AI generated answers in Lithuanian was conducted, producing a dataset with student answers, GPT generated answers and its paraphrased versions. The authors~\cite{weegar2024reducing} created a dataset of student answers in Swedish in programming languages, networking and the Internet, and data abstractions and manipulations. The authors examined different machine learning methods to tackle the task. In~\cite{klevstuen2022assisting}, the use of information retrieval and text mining methods were investigated to evaluate the content of Norwegian exam answers in Computer Science. In our work, we release multi-domain publicly available datasets as well as benchmark results for some of the open-source multilingual LLMs.

\section{Datasets}

To generate answer matching datasets, the three-stage pipeline was implemented.

Firstly, we used the approach for generating question-answer Knowledge and Instruction Dataset (KID) based on Wikipedia, introduced in~\cite{kiulian2024bytes} and adapted it for Latvian (Lat-KID) and Lithuanian (Lit-KID). More details are provided in the Section~\ref{sec:kid}. The generated datasets consist of pairs of a question and a reference answer (assumed to be correct and relevant to the question), as well as a factual information that supports the answer. 

On the second stage, for each pair of question and answer, we defined a list of different \textit{alteration rules} that rewrites reference answer to matched or non-matched (more details are provided in the Section~\ref{sec:answers_gen_process}). We used GPT-4o and LLaMa3 8b (see Figure~\ref{fig:ds_example}), utilizing separate prompts for each rule. The non-matched prompts were composed in a way that preserves as much words and semantics of the reference answer as possible with changing key words of the answer, while matched prompts are more flexible. 

Finally, the generated results were validated and methods were filtered based on the accept ratio (more details are provided in the Section~\ref{sec:answer_gen_eval_man}).

\begin{figure}
    \centering
    \includegraphics[width=\linewidth]{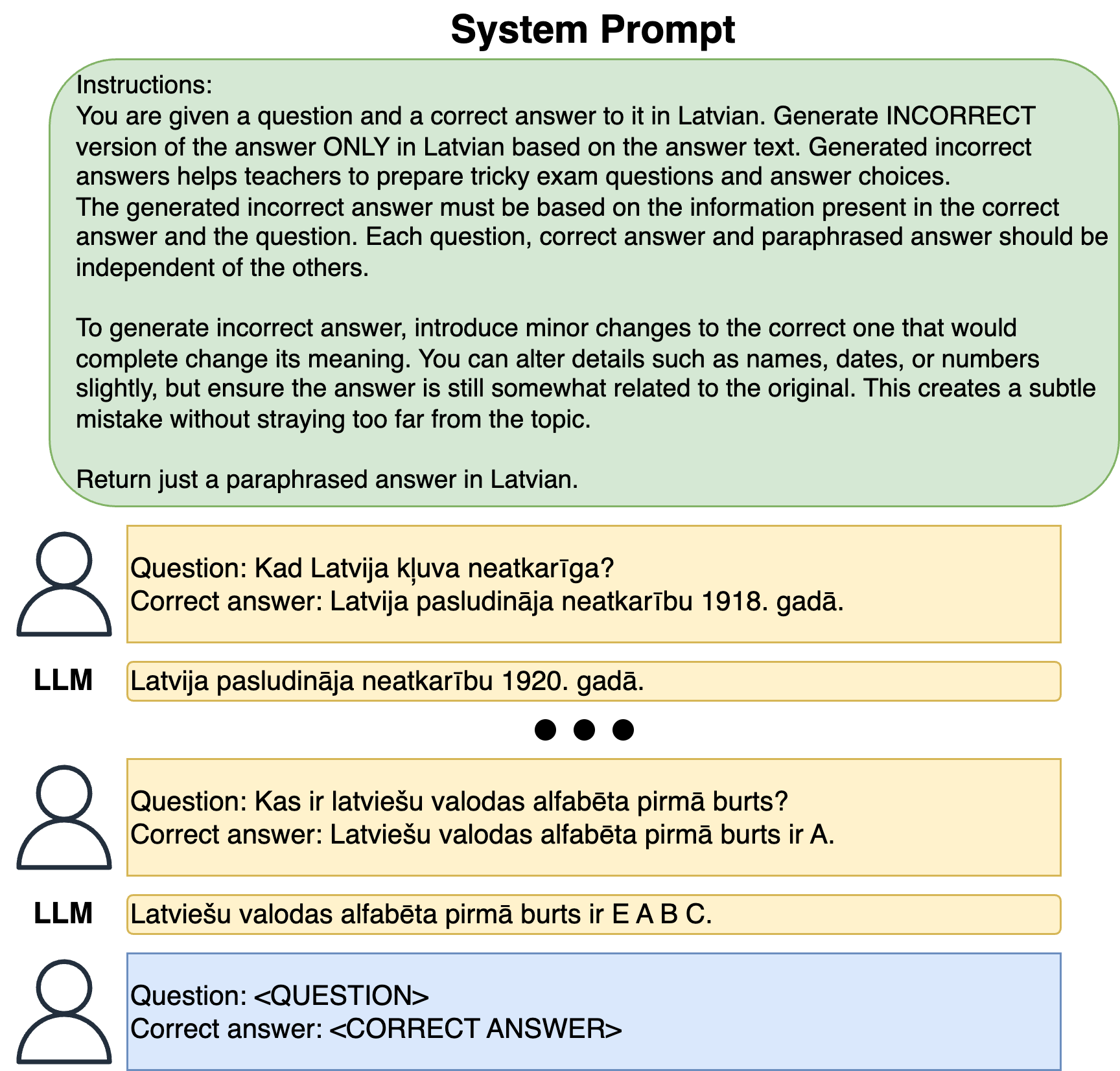}
    \caption{Example of few-shot incorporating minor changes prompt for non-matched answers generation in Latvian. $<\dots>$ indicate the sample that requires prediction.}
    \label{fig:na_prompts}
\end{figure}

\subsection{Lat-KID and Lit-KID Question-Answering Datasets} \label{sec:kid}
For each language, we extracted top 1,000 articles for each month of the last 12 month from Wikipedia, resulting in 12,000 articles. From this pool, 1,000 articles with the top cumulative counts were extracted. The articles were filtered by their relevance to the corresponding country with Gemini 1.5 Pro~\cite{geminiteam2024gemini15unlockingmultimodal}. Each article was separated into the paragraphs and at least 3 questions were generated for it with Gemini 1.5 Pro. The prompt contains additional fields to run a self-check on the quality of the question (standalone, in the correct language, natural sounding). The prompts are available in the project's GitHub.

The obtained Lat-KID dataset has 502 unique questions. The average number of words in the question is 9.83 and in the answer is 24.37. The total number of words in the dataset (questions and reference answers) is 17,172. The unique amount of words is 5,058.

The obtained Lit-KID dataset has 690 unique questions. The average number of words in the question is 9.88 and in the answer is 29.02. The total number of words in the dataset (questions and reference answers) is 26,849. The unique amount of words is 7,725.

\subsection{Matched and Non-Matched Answers Generation} \label{sec:answers_gen_process}

\paragraph{Non-Matched Answers Generation.}

We defined two alteration rules for non-matched answers generation: incorporating minor changes (\textbf{IMC}) and changing domain related information (\textbf{CDRI}). IMC includes changes to the text that change a couple of key words like date, name, location etc, while keeping everything else unchanged. CDRI is similar to IMC, however its objective is to change a key term to the similar \textit{from the same domain}. For example, changing the name of the first president to the second one, etc. 

To generate non-matched answers, we utilized LLaMa3 7b\footnote{After manual evaluation, only IMC were generation was accepted for Lat-KID and CDRI for Lit-KID.} and GPT-4o\footnote{We experimented with LLaMa2 13b, however manual evaluation showed much worse results.}.

When generating IMC and CDRI answers, the model was presented with the few-shot example prompts (see Figure~\ref{fig:na_prompts}).

\paragraph{Matched Answers Generation.}
We defined the following alteration rules for matched answers generation: adding more question-related entities (\textbf{Ents}), changing words to synonyms (\textbf{Synonyms}), adding more background information (\textbf{MoreInfo}), and style swap to exclamatory (\textbf{Exclamatory}).

As previously, we used GPT-4o and LLaMa3 7b. The models were presented with different prompts per rule. The code and prompts are available in the GitHub repository for the project\footnote{HIDDEN FOR REVIEW} .

\paragraph{Postprocessing.} After the generating answers, the duplicates were removed. The resulting amount of (question, reference answer, generated answer) triplets is 3,012 (1,506 are matched and other 1,506 are non-matched) for Latvian and 4,830 (2,760 are matched and 2,070 are non-matched) for Lithuanian. The amount of matched answers is 3,697. The amount of non-matched answers is 1,809. 

\subsection{Manual Evaluation} \label{sec:answer_gen_eval_man}

We recruited two native speakers for Latvian and Lithuanian to evaluate the quality of the final generated dataset. They were presented with a random triplet of (question, reference answer, generated answer) and a description if the generated answer was generated by matched or non-matched method. Based on that, the annotators had to accept a triplet if the description fits the reference and generated answers. Otherwise, they had to reject sample. The results are presented in Appendix~\ref{sec:annots}.

\section{Methodology}

To evaluate the LLMs capabilities and an influence of the prompting strategy, we used two prompting methods per language for this task: zero shot (\textbf{ZS}) and few shot (\textbf{FS}). We set all the parameters to defaults with a random seed of 2.

In all of the methods, the model were instructed to start their output with \textit{True} if the provided reference answer and a generated answer are matched otherwise with \textit{False}. ZS and FS shared the same system prompt, but FS gave a model additional examples in corresponding language.

We evaluated LLaMa3.1 (8b and 70b)~\citep{dubey2024llama3herdmodels}, Mistral Nemo 12b and Mistral 7b~\citep{jiang2023mistral7b}, and QWEN2.5 (7b and 72b)~\cite{qwen2.5, qwen2} models. To parse the output, we checked if the model followed instructions about the output. If it did not, we retrieved the key words: ``True'' or ``False''. If none of the words were presented, we counted it as an incorrect prediction.

\section{Results and Discussion}
\begin{table}
\begin{tabular}{|l|ll|ll|}
\hline
                          & \multicolumn{2}{c|}{\textit{LT}}                    & \multicolumn{2}{c|}{\textit{LV}}                    \\ \hline
                          & \multicolumn{1}{c|}{ZS}   & \multicolumn{1}{c|}{FS} & \multicolumn{1}{c|}{ZS}   & \multicolumn{1}{c|}{FS} \\ \hline
\textbf{QWEN2.5 72b}      & \multicolumn{1}{l|}{0.99} & 0.99                    & \multicolumn{1}{l|}{0.99} & 0.99                    \\ \hline
\textbf{LLaMa3.1 70b}     & \multicolumn{1}{l|}{0.99} & 0.99                    & \multicolumn{1}{l|}{0.99} & 0.99                    \\ \hline\hline
\textbf{Mistral Nemo 12b} & \multicolumn{1}{l|}{0.96} & 0.94                    & \multicolumn{1}{l|}{0.96} & 0.94                    \\ \hline\hline
\textbf{EuroLLM 9b}       & \multicolumn{1}{l|}{0.13} & 0.97                   & \multicolumn{1}{l|}{0.05} &  0.84                   \\ \hline
\textbf{LLaMa3.1 8b}      & \multicolumn{1}{l|}{0.89} & 0.98                    & \multicolumn{1}{l|}{0.87} & 0.96                    \\ \hline
\textbf{QWEN2.5 7b}       & \multicolumn{1}{l|}{0.98} & 0.98                    & \multicolumn{1}{l|}{0.97} & 0.97                    \\ \hline
\textbf{Mistral 7b}       & \multicolumn{1}{l|}{0.95} & 0.91                   & \multicolumn{1}{l|}{0.95} & 0.91                   \\ \hline

\end{tabular}
\caption{F1 scores of binary matching. \textit{LT} and \textit{LV} refer to Lithuanian and Latvian respectively. ZS and FS refer to zero shot and few shot respectively.} \label{tab:results}

\end{table}

\begin{figure*}
\begin{minipage}{.5\textwidth}
    \subfloat[Zero shot]{\includegraphics[width=\textwidth]{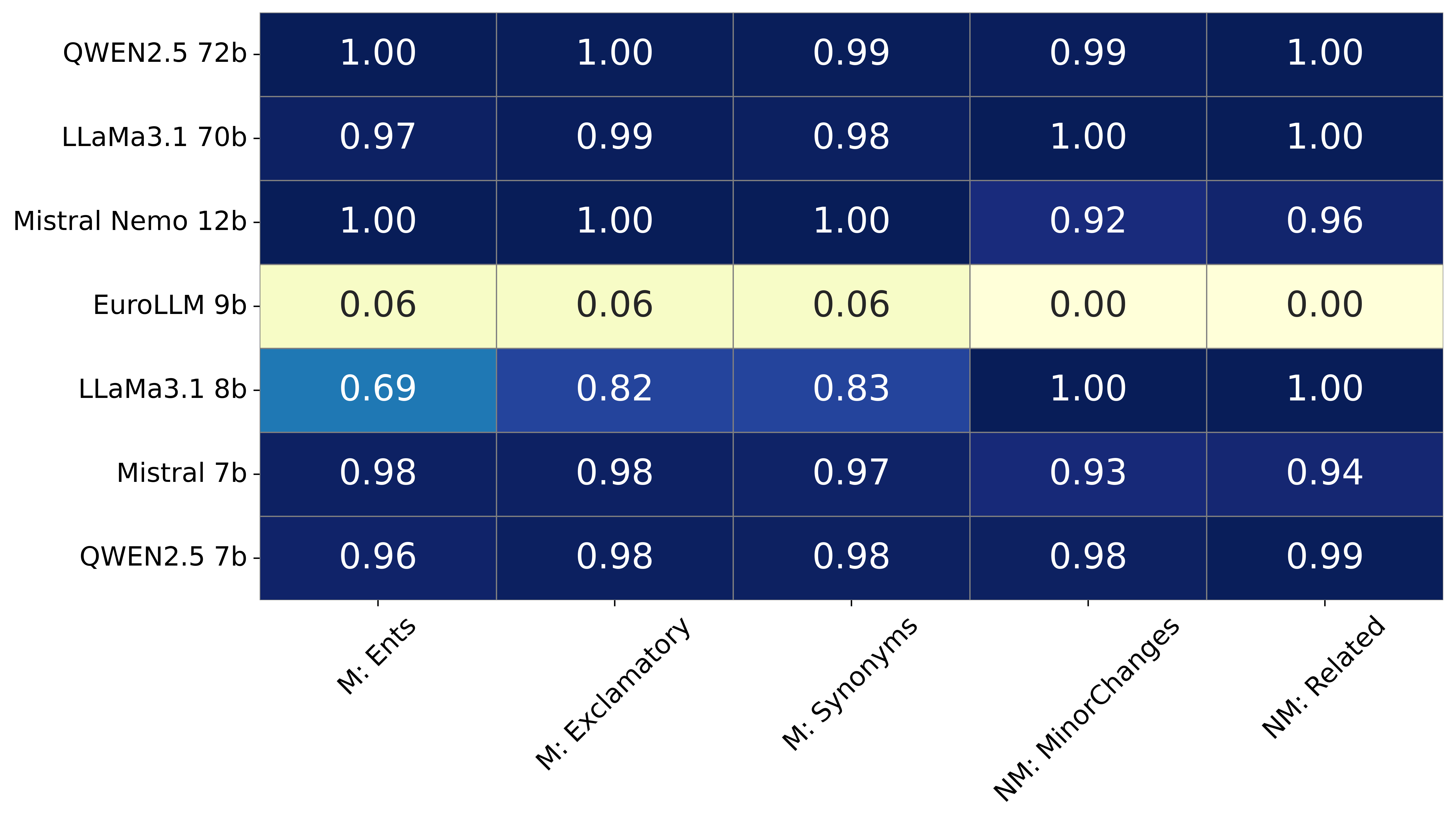}}
\end{minipage}
\hfill    
\begin{minipage}{.5\textwidth}
    \subfloat[Few shot]{\includegraphics[width=\textwidth]{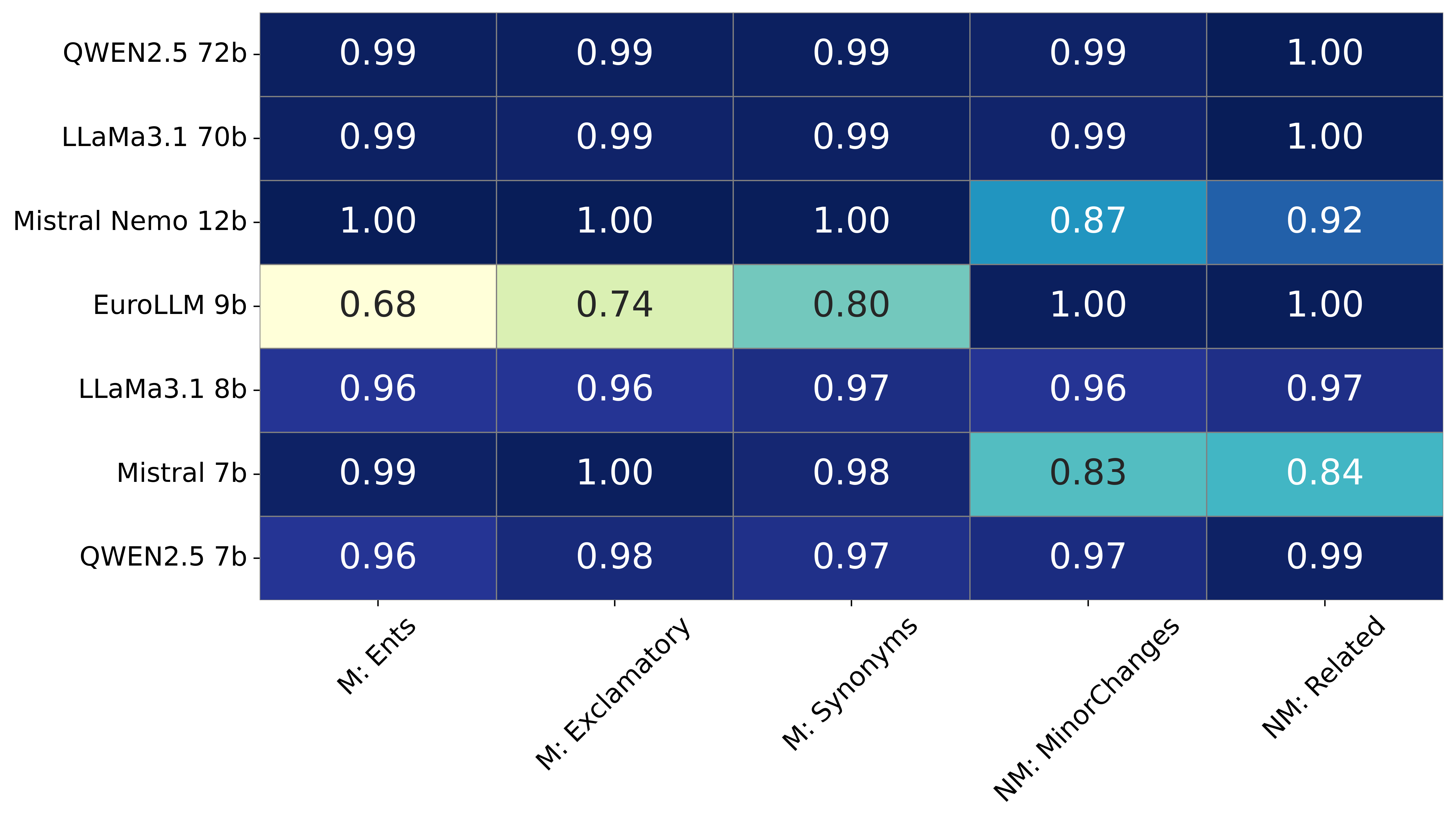}}
\end{minipage}
    \caption{Accuracy scores per generated answer type for Latvian.}\label{fig:lav_res}
\end{figure*}

\begin{figure*}
\begin{minipage}{.5\textwidth}
    \subfloat[Zero shot]{\includegraphics[width=\textwidth]{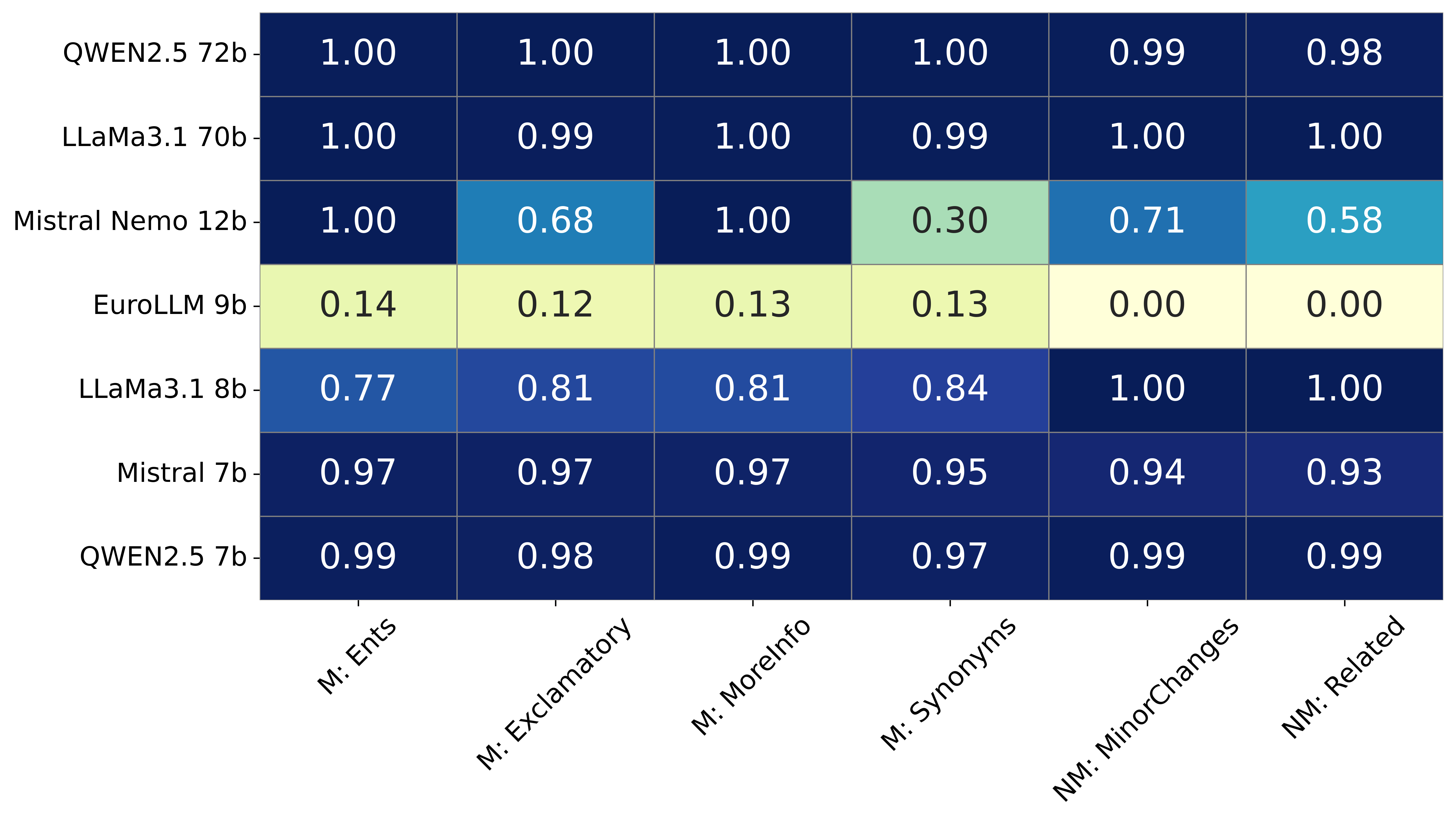}}
\end{minipage}
\hfill    
\begin{minipage}{.5\textwidth}
    \subfloat[Few shot]{\includegraphics[width=\textwidth]{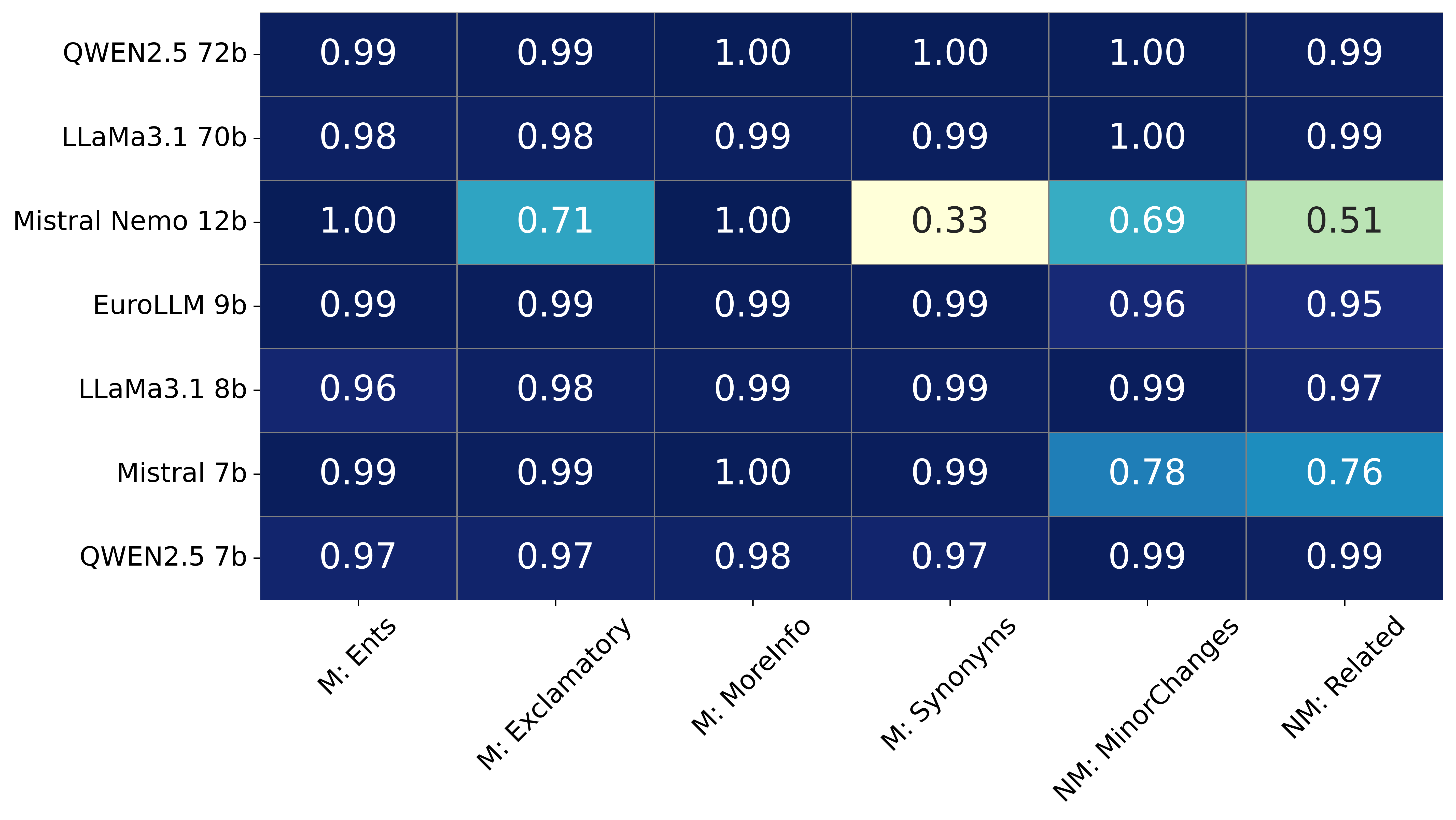}}
\end{minipage}
    \caption{Accuracy scores per generated answer type for Lithuanian.}\label{fig:lt_res}
\end{figure*}

The results are presented in Table~\ref{tab:results}, and on Figures~\ref{fig:lav_res} and~\ref{fig:lt_res}. Additionally, we measured a percentage of times, when model followed the provided format and started with ``True'' or ``False''. The majority of models were able to output the correct format for 99\% on Latvian samples.  For Lithuanian, LLaMa3.1 8b generated text in correct format in 89\% of times in ZS settings. In case of the FS, this value is 99\%. Other models consistently followed the format with a rate of 99\%. EuroLLM 9b was not able to follow a format at all in ZS settings for both languages, even though its results were legible, but impossible to parse. However, when presented with a few shot examples, it generated expected format.

Our results demonstrated that larger LLMs (with 70b parameters) are capable of reliably detect matched and non-matched answers in Lithuanian and Latvian. We hypothesized that LLMs would output near perfect scores, however, smaller models performed differently. In case of Mistral Nemo, there was a slight decrease of results when switched from zero shot to a few shot approach in both languages. On the contrary, LLaMa3.1 8b performed better in a few shot scenario, improving its ZS score on 9\%. QWEN2.5 7b performed nearly perfectly, achieving 99 accuracy score in both settings.

Deeper analysis of results indicated that in case of Latvian, most of the models (except for LLaMa3.1 8b, MIstral 7b, and EuroLLM 9b) showed almost perfect performance on all the generated types of matched and non-matched answers. LLaMa3.1 8b was able to pick up non-matched answers in ZS and FS settings, but struggled with matched answers, demonstrating a bias towards negative answers. However, exposing it with the additional examples boosted its scores to the same level as others. EuroLLM was not able to follow instructions in zero shot prompts, therefore performing poorly. However, in the few shot settings, the model was able to detect non-matched answers, but had less success with matching answers, demonstrating bias towards negative answers. Mistral 7b perfromed well in ZS experiments, but showed a weaker performance in FS for non-match generated samples. 

For Lithuanian, the least reliable model was Mistral Nemo 12b. It demonstrated a strong performance on the matched answers with more information and more entities, but was not able to effectively detect synonyms changes in both ZS and FS settings. In case of this model, providing more examples to the model did not have a noticeable effect. Interestingly, EuroLLM showed the same pattern as for Latvian in ZS, but was able to get a comparable results with the 70b groups of models in FS settings. It indicates that the model has a better understanding of Lithuanian than Latvian when it comes to this task, and can perform well when provided with examples. 

Therefore, based on our observations, we can address each of the research questions we formulated. 

\textbf{Q1:} Are LLMs capable of correctly identifying matched and non-matched answers with the proposed alteration rules ? Overall, the evaluated models were able to accurately identify, which answers are matched and which are not. LLMs with the greater number of parameters showed a very consistent performance, when smaller model can have difficulties with Latvian or Lithuanian. Specifically, LLaMa3.1 8b and EuroLLM 9b require additional examples, when QWEN2.5 7b and Mistral 7b are on par with the larger models. Moreover, we found specific types of alternation rules that models had more difficulties to pick up. Specifically LLaMa3.1 8b and EuroLLM 9b had difficulties with added entities in the text in Latvian. Mistral 7b struggled with incorporating minor changes  and changing domain related information rules in Latvian FS settings. Mistral Nemo obtained weaker performance on changing words to synonyms and style swap to exclamatory (\textbf{Exclamatory}) rules in Lithuanian.

\textbf{Q2:} Is there a difference between few-shot and zero-shot inference for different LLMs for this task ? Our findings showed that few shot approach did not improve the scores of the larger models: they are already very high. However, it can be helpful in case of some smaller models, especially with EuroLLM 9b. In case of Mistral 7b, the perfromance was decreased with adding more examples. On the other hand, if the model struggles with a language, providing more examples will not necessarily improves its performance (e.g. Mistral Nemo in Lithuanian or Mistral 7b) for this task.

\section{Conclusion}

In conclusion, our findings demonstrate that large language models (LLMs) with greater parameter counts, such as QWEN2.5 72b and LLaMa3.1 70b, consistently achieve high accuracy in distinguishing matched and non-matched answers across both Latvian and Lithuanian, regardless of zero-shot or few-shot settings. Smaller models showed less robustness, with LLaMa3.1 8b and EuroLLM 9b benefiting from additional examples in few-shot scenarios. Mistral Nemo 12b struggled with detecting certain nuances, particularly in Lithuanian. QWEN2.5 7b and Mistral 7b were able to obtain a similar the performance to the larger 70b models, but in case of Mistral 7b the performance decreased in with a few shot approach. These results highlight the robustness of larger models and the potential for targeted improvements in smaller ones to address answer matching task with the defined set of alteration rules.



\bibliographystyle{acl_natbib}
\bibliography{nodalida2025}

\appendix

\begin{table*}
    \centering
    \begin{tabular}{|c|c|c|c|c|c|c|}

    \hline
        \textbf{Lit R} & \textbf{Lit A} &  \textbf{Lat R} & \textbf{Lat A} & \textbf{Model} & \textbf{Class} \\ \hline

        1 & 29 & 1 & 29 & GPT-4o & Match Ents \\ \hline
        0 & 30 & 4 & 26 & GPT-4o & Match MoreInfo \\ \hline
        1 & 29 & 2 & 28 & GPT-4o & Match Syns \\ \hline
        0 & 30 & 2 & 28 & GPT-4o & Match Style \\ \hline
        0 & 60 & 3 & 56 & GPT-4o & Non-Match MinorCh. \\ \hline
        3 & 57 & 2 & 57 & GPT-4o & Non-Match Relat. \\ \hline\hline

        20 & 10 & 12 & 18 & LLaMa2:13b & Match Ents \\ \hline
        22 & 8 & 15 & 14 & LLaMa2:13b & Match MoreInfo \\ \hline
        13 & 17 & 10 & 19 & LLaMa2:13b & Match Syns \\ \hline
        12 & 18 & 14 & 16 & LLaMa2:13b & Match Style \\ \hline
        16 & 43 & 14 & 15 & LLaMa2:13b & Non-Match MinorCh. \\ \hline
        12 & 46 & 13 & 32 & LLaMa2:13b & Non-Match Relat. \\ \hline\hline

        10 & 20 & 5 & 25 & LLaMa3:7b & Match Ents \\ \hline
        5 & 25 & 8 & 22 & LLaMa3:7b & Match MoreInfo \\ \hline
        8 & 22 & 3 & 26 & LLaMa3:7b & Match Syns \\ \hline
        13 & 16 & 14 & 16 & LLaMa3:7b & Match Style \\ \hline
        6 & 54 & 4 & 56 & LLaMa3:7b & Non-Match MinorCh. \\ \hline
        2 & 58 & 6 & 48 & LLaMa3:7b & Non-Match Relat. \\ \hline\hline
        
        5 & 235 & 14 & 224 & GPT-4o & All \\ \hline
        95 & 142 & 79 & 144 & LLaMa2:13b & All \\ \hline
        44 & 195 & 40 & 193 & LLaMa3:7b & All \\ \hline
 
    \end{tabular}
    \caption{Annotation results. \textbf{R} and \textbf{A} indicate amount of rejected and accepted samples respectively with the language at the beginning. \textbf{Class} indicates a generation prompt that was used and whether it should match with the reference answer.}
    \label{tab:annot_res}
\end{table*}

\section{Manual Evaluation} \label{sec:annots}

For each language, we recruited two native speakers to evaluate the outputs of LLMs on the answers generation task. Each annotator was presented with 360 random samples from the dataset. Each sample contained a question, a reference answer, a generated answer with an instruction on whether it supposed to be matched with the reference answer. If the reference answer and the generated answer are matched and they are supposed to be matched or the reference answer and the generated answer are not matched and they are not supposed to be matched, the label \textit{accept} was assigned to the sample. Otherwise, the label \textit{reject} was assigned. For each model (LLaMa2:13b, GPT-4o, and LLaMa3:7b) and for each matched generation type, the annotators were presented with 25 samples. For non-matched generation methods, the annotators were presented with 40 samples. The aggregated results (after cleaning the duplicates) are presented in the Table~\ref{tab:annot_res}

Based on the the results, we kept LLaMa3 Non-Match Relat. generation results and all of the GPT-4o generated results in the dataset for Lithuanian. Similarly, we kept LLaMa3 Non-Match MinorChanges and GPT-4o (except for Match MoreInfo, which was excluded by mistake) in the dataset for Latvian. Our results indicate that GPT-4o is capable of generating matched and non-matched answers with different methods in these languages, when LLaMa3 and LLaMa2 struggle. 
\end{document}